\documentclass[sigconf]{acmart}
\usepackage{multirow}
\usepackage{subfig}
\usepackage[group-separator={,}]{siunitx}
\AtBeginDocument{%
  \providecommand\BibTeX{{%
    \normalfont B\kern-0.5em{\scshape i\kern-0.25em b}\kern-0.8em\TeX}}}

\copyrightyear{2023}
\acmYear{2023}
\setcopyright{acmlicensed}
\acmConference[MM '23] {Proceedings of the 31st ACM International Conference on Multimedia}{October 29--November 3, 2023}{Ottawa, ON, Canada.}
\acmBooktitle{Proceedings of the 31st ACM International Conference on Multimedia (MM '23), October 29--November 3, 2023, Ottawa, ON, Canada}
\acmPrice{15.00}
\acmISBN{979-8-4007-0108-5/23/10}
\acmDOI{10.1145/3581783.3612179}
\acmSubmissionID{2129}

\begin{document}

\title{Text-Only Training for Visual Storytelling}


\author{Yuechen Wang$^1$, \quad Wengang Zhou$^{1,2}$$^{\dagger}$, \quad Zhenbo Lu$^2$$^{\dagger}$, \quad Houqiang Li$^{1,2}$}
\makeatletter
\def\authornotetext#1{
	\if@ACM@anonymous\else
	\g@addto@macro\@authornotes{
		\stepcounter{footnote}\footnotetext{#1}}
	\fi}
\makeatother
\authornotetext{Corresponding authors: Wengang Zhou and Zhenbo Lu.}

\affiliation{%
	\institution{$^1$CAS Key Laboratory of Technology in GIPAS, EEIS Department, \\ University of Science and Technology of China}
	\country{}
}
\affiliation{%
	\institution{$^2$Institute of Artificial Intelligence, Hefei Comprehensive National Science Center}
	\country{}
}
\email{ wyc9725@mail.ustc.edu.cn, zhwg@ustc.edu.cn, luzhenbo@iai.ustc.edu.cn, lihq@ustc.edu.cn}


\renewcommand{\shortauthors}{Yuechen Wang, Wengang Zhou, Zhenbo Lu, \& Houqiang Li}

\begin{abstract}
Visual storytelling aims to generate a narrative based on a sequence of images, necessitating both vision-language alignment and coherent story generation. Most existing solutions predominantly depend on paired image-text training data, which can be costly to collect and challenging to scale. To address this, we formulate visual storytelling as a visual-conditioned story generation problem and propose a text-only training method that separates the learning of cross-modality alignment and story generation.
Our approach specifically leverages the cross-modality pre-trained CLIP model to integrate visual control into a story generator, trained exclusively on text data. Moreover, we devise a training-free visual condition planner that accounts for the temporal structure of the input image sequence while balancing global and local visual content. The distinctive advantage of requiring only text data for training enables our method to learn from external text story data, enhancing the generalization capability of visual storytelling.
We conduct extensive experiments on the VIST benchmark, showcasing the effectiveness of our approach in both in-domain and cross-domain settings. Further evaluations on expression diversity and human assessment underscore the superiority of our method in terms of informativeness and robustness.

\end{abstract}




\maketitle

\section{Introduction}

\begin{figure}[tb]
    \centering
    \includegraphics[scale=0.47]{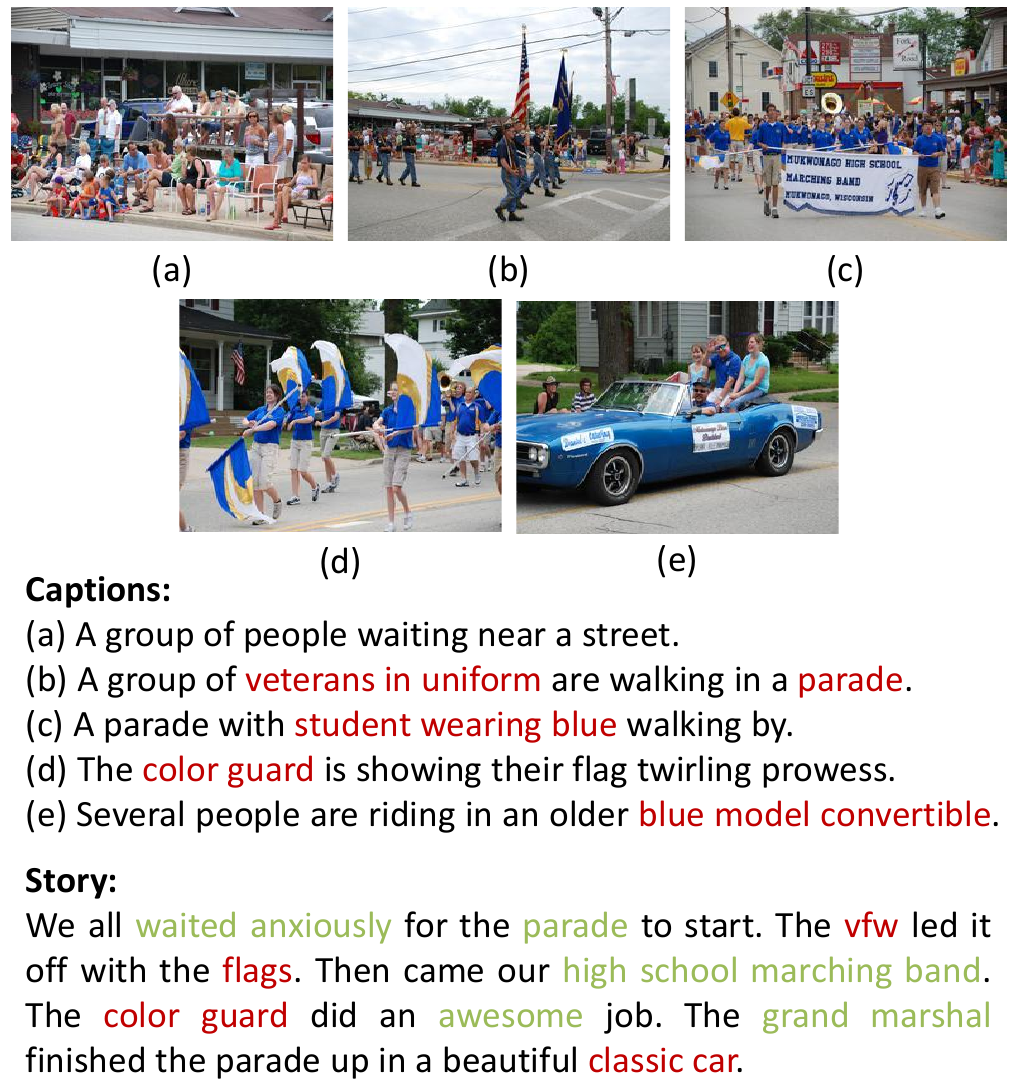}
    \caption{Example difference between image captioning and visual storytelling.
    Words highlighted in \textcolor[RGB]{192,0,0}{red} represents contents in the corresponding image, and words highlighted in \textcolor[RGB]{157,187,97}{green} represents subjective expressions and information reasoned from other images.
    }
    \label{fig:fig1}
\vspace{-1.5em}
\end{figure}

Visual storytelling~\cite{Huang2016VisualS}, a task aimed at generating narratives based on image sequences, has received significant interest due to its potential applications in diverse domains such as advertising, entertainment, and education. In comparison to other vision-to-language generation tasks, such as visual captioning~\cite{Devlin2015LanguageMF}, visual storytelling presents unique challenges stemming from its subjective and imaginative nature.
As illustrated in Fig.~\ref{fig:fig1}, to create a coherent story that aligns with the visual input, each sentence must not only describe the corresponding image, but also maintain logical connections to both preceding and subsequent sentences. This dual requirement of ensuring cross-modality consistency while preserving narrative coherence constitutes the primary challenge of visual storytelling.

Existing works often require large amounts of labeled data and attempt to learn both cross-modality alignment and story coherence simultaneously through end-to-end training. By training on large manually annotated data, these models are capable of generating coherent and visual-related stories.
Subsequent advancements, such as the incorporation of external knowledge~\cite{ijcai2019p744} and scene graphs~\cite{Wang2020StorytellingFA}, have further enriched generated stories with additional details. 
More recently, the employment of large pre-trained Transformer-based language models has led to considerable improvements in visual storytelling~\cite{9577852}.
Nevertheless, the substantial cost associated with annotating and training extensive datasets remains a significant bottleneck, limiting the scalability of visual storytelling approaches.

On the other hand, the burgeoning capabilities of pre-trained models offer potential for leveraging these models to transfer knowledge to downstream tasks such as visual storytelling, facilitating more data-efficient learning.
To this end, some prior works have combined generative language models~\cite{Radford2018ImprovingLU, radford2019language, NEURIPS2020_1457c0d6} with cross-modality pretrained models~\cite{Radford2021LearningTV} to explore text-only training for image captioning~\cite{Tewel_2022_CVPR,nukrai-etal-2022-text}. 
However, while these cross-modality models trained on paired image-text data successfully align text with individual images, they are limited in their capacity to comprehend the temporal structure of image sequences—an essential component of visual storytelling.

\begin{figure*}[htb]
    \centering
    \includegraphics[scale=0.52]{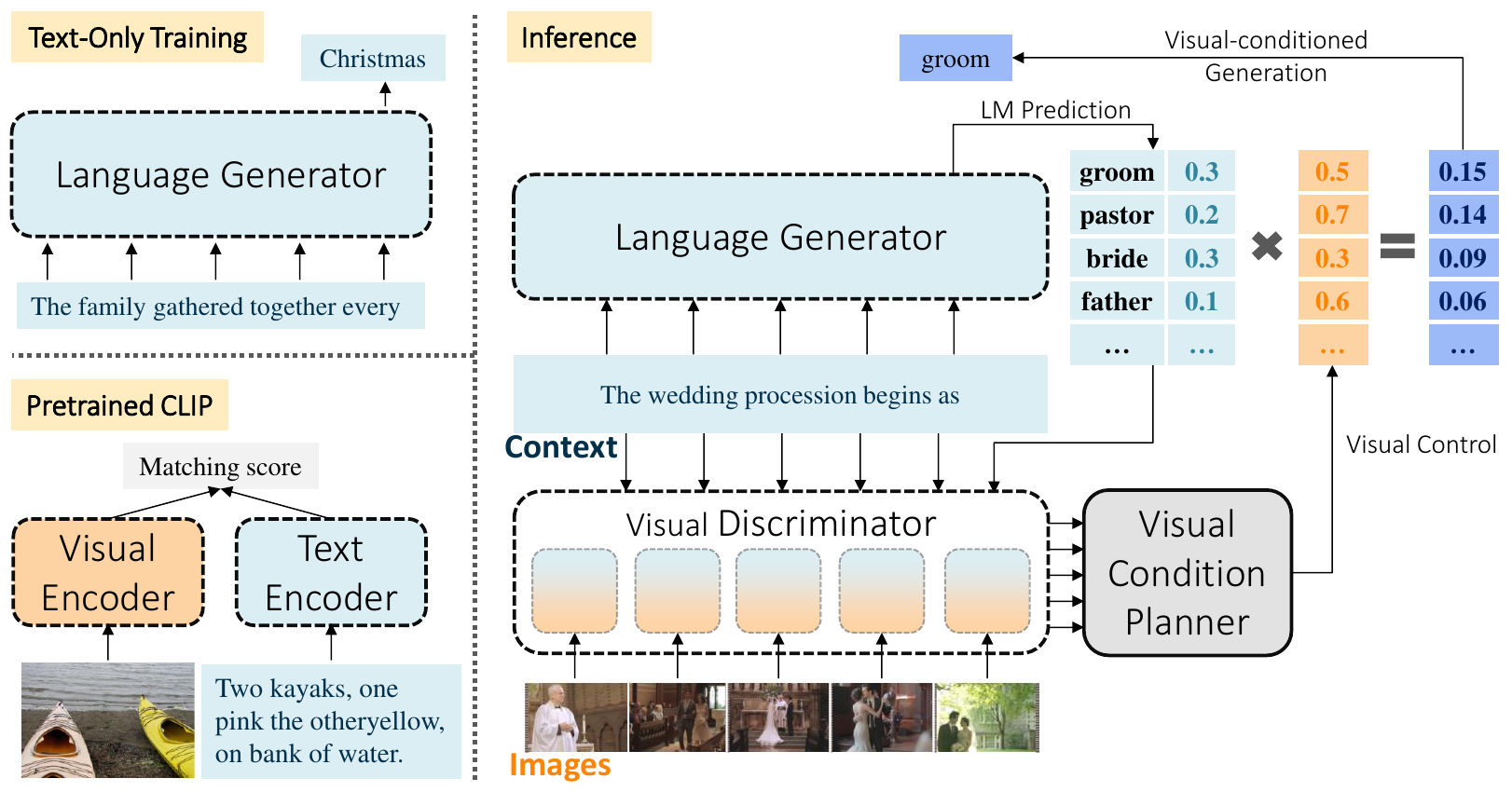}
    \caption{The training and inference pipeline of our method. 
    During training, we only train the language generator on a story dataset without visual information.
    Then, at inference time, we utilize a pretrained CLIP model as a visual discriminator to align images with candidate tokens. Additionally, we introduce a visual condition planner that aggregates image sequences, and the output visual control is then incorporated into the generation process.
    }
    \label{fig:pipeline}
\end{figure*}

Motivated by the observations discussed above, we propose a novel framework that leverages pretrained generative language models and cross-modality models for data-efficient visual storytelling. We formulate visual storytelling as a visual-conditioned story generation task. As shown in Fig.~\ref{fig:pipeline}, we first fine-tune a pretrained language model using only textual data to develop a story generator. Then, we incorporate visual clues during the generation process. 
Specifically, at each decoding step, we use a pretrained cross-modality model CLIP~\cite{Radford2021LearningTV} as a visual discriminator to compute a matching score between candidate text and input images. 
A visual condition planner is then designed to aggregate the matching results of the input images, emphasizing semantics in the corresponding image while retaining information from other images. 
Finally, the aggregated result is incorporated into the decoding probability distribution to guide the generation of the next token, resulting in a coherent and visually aligned story.

To demonstrate the effectiveness of our proposed method, we conduct extensive experiments on the widely-used VIST benchmark~\cite{Huang2016VisualS}. 
The results show that our approach achieves state-of-the-art performance on various evaluation metrics including comparing-based automatic metrics, statistics-based metrics, and human evaluation. 
Additionally, our method exhibits impressive generalization ability in domain-transfer experiments, suggesting its potential for real-world applications.

We summarize the major contributions of this work as follows:
\begin{itemize}
    \item 
    We formulate visual storytelling as a visual-conditioned generation problem and propose a data-efficient framework which is trained solely on text-only data by leveraging pretrained CLIP model.
    
    \item We introduce a visual condition planner which is free of training. The planner aggregates sequential visual inputs to provide local details while maintaining the global theme of the image album, thereby improving the quality of generated stories.
    
    \item Extensive experiments on VIST benchmark demonstrate the effectiveness of our proposed method, as evidenced by its superior performance compared to existing methods in both automatic metrics and human evaluations.
\end{itemize}

\section{Related Work}
The main idea of our work is to model visual storytelling as a controlled text generation task, and exploit large pretrained models to reduce the cost of cross-modality training. In this section, we provide a brief review of the related areas.

\subsection{Visual Storytelling}
Visual storytelling was first introduced by Huang et al.\cite{Huang2016VisualS}, which involves the use of a sequence of images to convey a narrative, necessitating reasoning over temporal context rather than merely understanding a static moment. Early approaches expanded upon conventional image captioning models by learning contextualized image representations\cite{Liu2017LetYP} and incorporating global visual information~\cite{GonzalezRico2018ContextualizeSA}. Additionally, reinforcement learning was employed to learn an implicit reward function through adversarial reward learning, optimizing the policy model to better align with human demonstrations~\cite{wang-etal-2018-metrics}. 
Hierarchical architectures~\cite{yu-etal-2017-hierarchically} and hierarchical reinforced training~\cite{Huang2018HierarchicallySR} have also demonstrated effectiveness in learning high-level semantics.

Given the imaginative nature of storytelling, external knowledge graphs have been integrated to introduce fictional concepts not present in images~\cite{ijcai2019p744,Hsu2019KnowledgeEnrichedVS,9996128}. To provide richer stories with greater visual detail, Wang et al.\cite{Wang2020StorytellingFA} incorporated scene graph generation, while Li et al.\cite{Li2019InformativeVS} learned cross-modal rules for mining visual concepts. Braude et al.\cite{Braude2021OrderedAF} proposed an ordered image attention approach to enhance story coherence through consistent grounding across sequenced images. Furthermore, Transformer-based frameworks have demonstrated capabilities in modeling spatial relationships between objects in images\cite{10.1145/3474085.3475236}.
In light of the proliferation of large pre-trained models, several studies have focused on leveraging pre-trained models (PTMs) for visual storytelling. Strategies include fine-tuning pre-trained Transformer encoders~\cite{10.1145/3487553.3524649,10.1145/3469877.3490604} and jointly tuning pre-trained language generation models with pre-trained image encoders~\cite{9577852}.

While the aforementioned approaches have demonstrated improvements in generated stories by incorporating external models, knowledge, and annotations, they also result in a significant increase in computational cost. In contrast, our proposed method circumvents the challenges associated with cross-modality training and annotation expenses by exclusively focusing on training using a text corpus.

\subsection{Controlled Text Generation}
In natural language generation, incorporating controllable constraints for open-ended text generation is both important and fundamental~\cite{prabhumoye-etal-2020-exploring}. With the advancements in pretraining, recent efforts have primarily concentrated on adapting pre-trained language models (LMs) to various attributes. A straightforward approach involves fine-tuning a pre-trained LM to generate text with specific attributes~\cite{nan-etal-2021-dart,ribeiro-etal-2021-structural,10.1145/3447548.3467418,carlsson-etal-2022-fine}. Alternatively, it is feasible to design new large LM architectures or retrain large conditioned LMs from scratch~\cite{Keskar2019CTRLAC,zhang-etal-2020-pointer,chan2021cocon}.
Recently, the exponentially increasing scale and capacity of pre-trained LMs have made it more viable and promising to fix pre-trained parameters and guide generation through post-processing. Dethathri \emph{et al.}\cite{Dathathri2019PlugAP} first proposed this paradigm as Plug-and-Play language models, wherein an attribute discriminator updates LM hidden states through back-propagation for attribute-controlled text generation. To reduce the computational cost associated with classifier-like discriminators ranking generated text, fine-tuned small LMs have been employed as generative discriminators to guide the generation of large pre-trained LMs\cite{krause-etal-2021-gedi-generative,liu-etal-2021-dexperts,yang-klein-2021-fudge,10.5555/3505520.3505528}. 
Pascual \emph{et al.}~\cite{pascual-etal-2021-plug-play} extended the plug-and-play method to keyword constraints and designed a distribution shifting strategy to augment the decoding probability of keywords. 

Guided decoding methods have demonstrated remarkable flexibility in accommodating various constraint types and hold considerable potential due to their independence from language models. In this work, we model visual storytelling as a visual-conditioned story generation task and propose a visual-linguistic discriminator to guide the generation process.

\subsection{Large Pretrained Models}
\textbf{Generative language models.} 
Taking advantage of the parallelism in the Transformer architecture~\cite{10.5555/3295222.3295349}, generative language models have shown a remarkable improvement in their capabilities in the past few years. 
These models can be broadly classified into two categories based on their network architecture: Decoder-Only models~\cite{Radford2018ImprovingLU, NEURIPS2020_1457c0d6} and Encoder-Decoder models~\cite{10.5555/3455716.3455856,lewis-etal-2020-bart}. Pretrained on large corpora, these models can effectively transfer to various language generation tasks, such as summarization, question answering, and story generation, with limited or even no supervised data.

\noindent\textbf{Cross-modality pretrained models.}
As the foundation of visual-language understanding, the idea to align the two modalities and learn a joint embedding space has been investigated extensively in the past decade~\cite{10.5555/2999792.2999849,10.1007/978-3-319-46478-7_5,8099701,8237711}.
In recent years, large cross-modality aligning models based on Transformers have gained considerable attention~\cite{Radford2021LearningTV,Li2022BLIPBL,Li2023BLIP2BL}. A representative work is CLIP~\cite{Radford2021LearningTV}, which trains two encoders for image and text inputs using a contrastive loss. With 400 million data pairs for training, CLIP has demonstrated remarkable zero-shot capabilities on multiple downstream tasks.

\section{Preliminaries}
A standard generative language model predicts the probability distribution of the next token based on previous inputs, which can be formulated as $P_{LM}(x_t|x_{<t})$. As a result, the probability of a text sequence $\boldsymbol{x} = \{x_1, \dots, x_T\}$ can be modeled as follows:
\begin{equation}
P_{LM}(\boldsymbol{x}) = \Pi_{t=1}^{T}P_{LM}(x_t|x_{<t}).
\end{equation}
In order to incorporate controls during the generation process, a constraint $c$ can be added to form a conditioned language model. This model generates the probability distribution of the next token based on the history inputs and the control constraint, and can be formulated as:
\begin{equation}
P(\boldsymbol{x}|c) = \Pi_{t=1}^{T}P(x_t|x_{<t}, c).
\end{equation}

Krause et al.~\cite{krause-etal-2021-gedi-generative} designed a generative discriminator to predict the probability that every candidate text sequence corresponds to the given constraint, which is given as:
\begin{equation}
    P_\theta(c|x_t,x_{<t}) = \frac{P(c)\Pi_{t=1}^T P(x_t|x_{<t},c)}{\sum_{c'\in \{c,\overline{c}\}}P(c')\Pi_{t=1}^T P(x_t|x_{<t},c')}
\end{equation}
where $\theta$ represents the learned parameters of the discriminator.
Then, based on the Bayes rule, the conditioned language model can be decoupled as:
\begin{equation}
    \label{Eq:decoupledprob}
    P(x_t|x_{<t}, c) \propto P_{LM}(x_t|x_{<t})P_\theta(c|x_t, x_{<t}).
\end{equation}

Therefore, each step of the generation process is implemented by combining an unconditioned language modeling $P_{LM}(x_t|x_{<t})$, and an attribute discriminator $P_\theta(c|x_t, x_{<t})$ with the guided decoding strategy as described in Eq.~\eqref{Eq:decoupledprob}. Here the discriminator is trained externally and can be easily used with any language generator in a plug-and-play manner.

\section{Method}

Given a sequence of images $\mathcal{I} = \{I_1, \dots, I_N\}$, a visual storytelling approach aims to generate a multi-sentence story $\boldsymbol{x}$ by predicting the probability $P(\boldsymbol{x}|\mathcal{I})$. 
To achieve this, we propose a framework that combines a text-only trained language generator, a pretrained visual discriminator, and a visual condition planner.
Fig.~\ref{fig:pipeline} illustrates the training and inference pipeline of our method. During the training phase, we fine-tune the language generator using story text, while in the inference phase, we employ the pre-trained visual discriminator and the visual condition planner to guide the generation process.

\subsection{Text-Only Training}
Compared to other supervised visual storytelling methods, our approach offers a notable advantage in that it requires training only on a text corpus, resulting in significant cost reductions in both training and annotation efforts.
Specifically, we fine-tune a Transformer decoder-based language model on a text story corpus to bridge the gap between pretraining on generic text and generating coherent stories.  
Given a narrative text sequence $\boldsymbol{x}=\{x_1, \dots, x_T\}$, the language model is fine-tuned by minimizing the maximum likelihood estimation~(MLE) loss:
\begin{equation}
    \mathcal{L}_{MLE} = -\frac{1}{T}\sum_{t=1}^T \log P_{LM}(x_t|x_{<t})
\end{equation}

Inspired by Su et al.~\cite{su2022a}, we incorporate an additional contrastive objective $\mathcal{L}_{CL}$ to encourage the generation of diverse and distinct expressions. The objective is defined as:
\begin{equation}
    \mathcal{L}_{CL} = \frac{1}{T(T-1)}\sum_{i=1}^T\sum_{j=1,j\neq i}^T\max(0, \epsilon - s(x_i, x_i) + s(x_i, x_j)),
\end{equation}
where $\epsilon$ is a predefined margin, and $s$ is the cosine similarity between tokens, defined by:
\begin{equation}
    \label{Eq:token_sim}
    s(x_i, x_j) = \frac{h_{x_i}^Th_{x_j}}{|h_{x_i}||h_{x_j}|}.
\end{equation}
The overall training objective of the language generator is the combination of the above two losses:
\begin{equation}
    \mathcal{L} = \mathcal{L}_{MLE} + \alpha\mathcal{L}_{CL},
\end{equation}
where $\alpha$ is a hyper-parameter to balance the loss items.

After fine-tuning on a text story corpus, the language generator is able to generate coherent stories in a style that is aligned with the training data.
However, since the generation process of the language generator is solely based on textual input, it may not take into account any visual content or the desired topic of the story. To address this, we introduce a visual discriminator and a visual condition planner to control the story topic and add details to the generated sentences.

\begin{figure}[tb]
    \centering
    \includegraphics[scale=0.34]{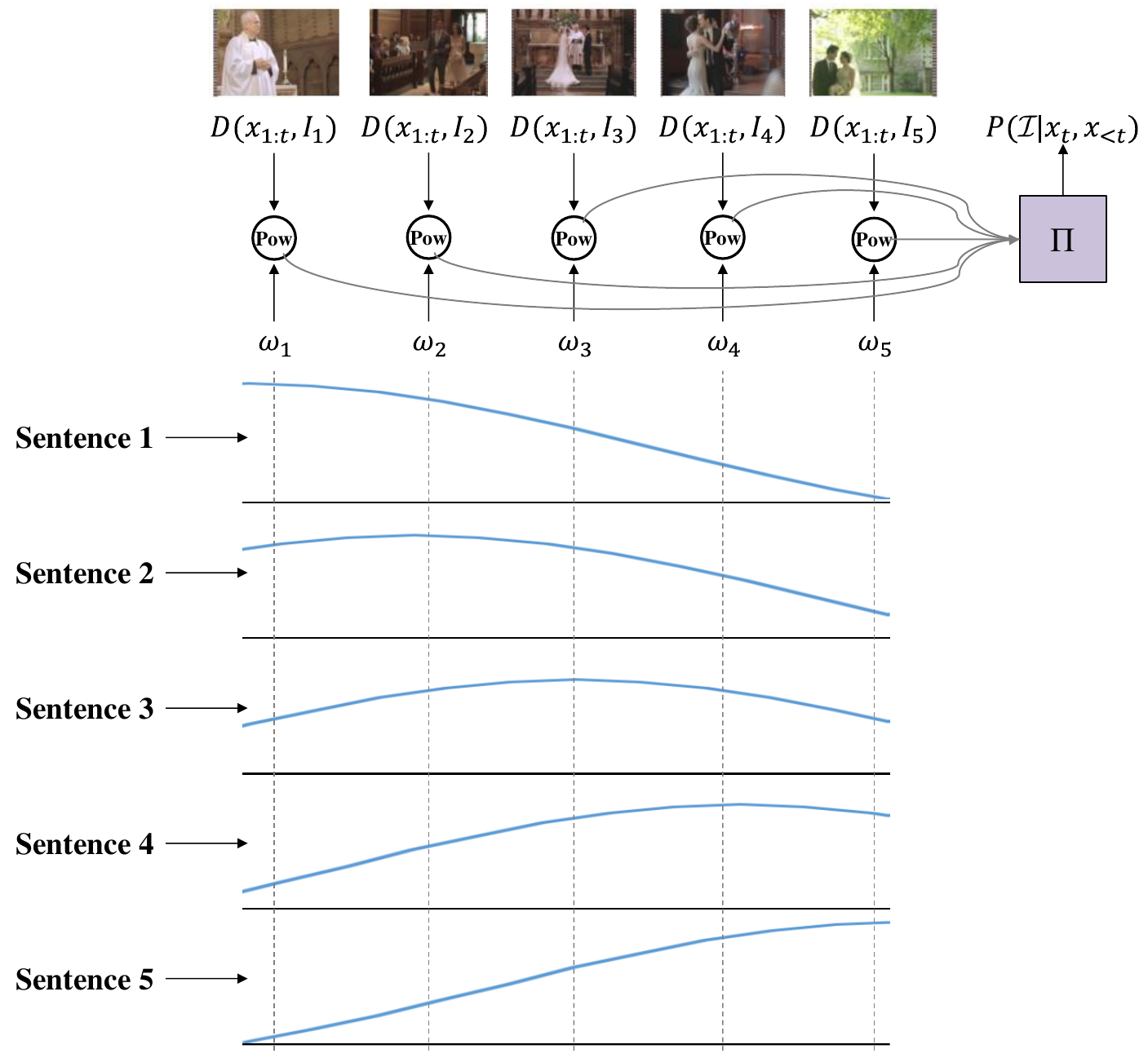}
    \caption{Illustration of the visual condition planner.}
    \label{fig:planner}
\vspace{-1.5em}
\end{figure}

\subsection{Visual Discriminator and Story Planning}

\begin{table*}[tb]
\renewcommand\arraystretch{1.2}
\begin{tabular}{ccccccccc}
\hline
\multicolumn{1}{c}{Method}                                              & METEOR  & BLEU-1 & BLEU-2 & BLEU-3 & BLEU-4 & ROUGE\_L & CIDEr \\ 
\hline
\multicolumn{8}{c}{\textit{Fully-Supervised Methods}} \\
\hline
INet~\cite{Jung2020HideandTellLT} & 35.6 & 64.4 & 40.1 & 23.9 & 14.7 & 29.7 & 10.0 \\
TAPM~\cite{9577852}               & 37.2 & -    & -    & -    & -    & 33.1 & 13.8 \\
OIAVist~\cite{Braude2021OrderedAF}& 36.8 & 68.4 & 42.7 & 25.2 & 15.3 & 30.2 & 10.1 \\
KAGS~\cite{9996128}               & 36.2 & 70.1 & 43.5 & 25.2 & 14.7 & 31.4 & 11.3 \\ 
\hline
\multicolumn{8}{c}{\textit{\textit{Text-Only Trained}}} \\  
\hline
Top-$k$ & 20.3 & 40.0 & 15.6 & 5.6 & 2.2 & 15.7 & 0.6 \\
Nucleus & 19.6 & 38.6 & 14.2 & 4.9 & 1.9 & 15.5 & 0.5 \\
MAGIC  & 20.3 & 41.2 & 16.1 & 5.9  & 2.8  & 16.0 & \textbf{1.3}  \\
\textbf{Ours}  & \textbf{23.0} & \textbf{43.7} & \textbf{20.2} & \textbf{9.2}  & \textbf{4.5}  & \textbf{17.3} & 1.2 \\ 
\hline
\end{tabular}
\caption{Comparison with existing methods on VIST test set. ``Fully-Supervised'' methods are trained on paired data, ``Text-Only Trained'' methods are trained on the textual stories of VIST. The best results under each metric are highlighted in bold.}
\label{Tab:mainresult}
\vspace{-1em}
\end{table*}

\begin{table*}[tb]
\renewcommand\arraystretch{1.2}
\begin{tabular}{ccccccccccccccc}
\hline
\multicolumn{1}{c}{\multirow{2}{*}{Method}} & \multicolumn{7}{c}{ROCStories} & \multicolumn{7}{c}{WritingPrompts} \\ 
\cmidrule(r){2-8} \cmidrule(r){9-15}
\multicolumn{1}{c}{}  & M  & B-1 & B-2 & B-3 & B-4 & R\_L & C  & M  & B-1 & B-2 & B-3 & B-4 & R\_L & C   \\ 
\hline
Top-$k$   & 15.3 & 28.6 & 9.5 & 2.5 & 0.7 & 12.1 & 0.2 & 15.0 & 26.8 & 8.0 & 2.0 & 0.4 & 12.2 & \textbf{0.2} \\
Nucleus & 15.0 & 28.4 & 9.0 & 2.4 & 0.7 & 12.0 & 0.3 & 14.3 & 25.6 & 7.3 & 1.5 & 0.3 & 11.9 & \textbf{0.2} \\
MAGIC       & 16.4 & \textbf{29.7} & 10.1 & 2.7 & 0.7 & 12.6 & 0.1 & 15.4 & 27.8 & 9.6  & \textbf{2.9} & 0.5 & 12.8 & \textbf{0.2} \\
\textbf{Ours}               & \textbf{16.6} & 28.6 & \textbf{11.5} & \textbf{3.8} & \textbf{1.2} & \textbf{12.9} & \textbf{0.2} & \textbf{16.2} & \textbf{28.8} & \textbf{9.9}  & \textbf{2.9} & \textbf{0.9} & \textbf{13.7} & \textbf{0.2} \\
\hline
\end{tabular}
\caption{Domain transfer results of text-only trained methods. The best results under each metric are highlighted in bold.}
\label{Tab:domsintrans}
\vspace{-2em}
\end{table*}

As previously mentioned, we consider visual storytelling as a visual-conditioned story generation task, and employ the guided decoding paradigm to integrate visual controls into the language generator.
To achieve this, we introduce a visual discriminator and a visual condition planner to score candidate sequences during generation. The visual discriminator is implemented using a pretrained visual-linguistic aligning model, while the visual condition planner is a training-free weighting model which aggregates the text matching results of different images.

During each generation step $t$, our language generator predicts a probability distribution $P_{LM}(x_t|x_{<t})$ over the vocabulary $V$ of possible next tokens, based on the context ${x_{<t}}$.
To guide the selection of candidate tokens, we employ a pretrained CLIP~\cite{Radford2021LearningTV} model as a visual discriminator $\mathbf{D}$.
Although the CLIP model has been pretrained on a large-scale dataset of paired visual and textual data, the pretraining process does not specifically involve annotations for visual storytelling. Therefore, by utilizing the pretrained CLIP, our method does not require cross-modality training and is capable of handling open-domain visual input. This makes our approach data-efficient and more scalable than previous methods.

Specifically, we feed each candidate token $x_t$ into the text encoder of CLIP along with the context tokens $x_{<t}$ to obtain a textual representation $f_{x_{1:t}}$. 
For each image $I_j$ in the input album, we extract a visual representation $f_{I_j}$ using the visual encoder of CLIP, where $j\in{1, \dots, N}$.
Then, the cosine similarity of $f_{x_{1:t}}$ and $f_{I_j}$ is computed as:
\begin{equation}
     \mathbf{D}(x_{1:t}, I_j) = \frac{f_{x_{1:t}}f_{I_j}}{|f_{x_{1:t}}||f_{I_j}|}, j\in\{1, \dots, N\}.
\end{equation}
As the CLIP model is trained to map visual and textual input representations into a sharing space, the matching score $\mathbf{D}(x_{1:t}, I_j)$ measures the relevance between candidate sequence $x_{1:t}$ and the input image $I_j$.

To ensure that the generated story aligns with the visual input fine-level semantics of the corresponding image to the sentence being generated and maintains the overall theme, we propose a visual condition planner. It aggregates the scores of the input images to derive a visual control for the current decoding step. Inspired by the work of Lin and Riedl~\cite{10.5555/3505520.3505528}, the planner does not require any training and achieves both global and local alignment through weighting and multiplication operations.

As depicted in Fig.~\ref{fig:planner}, the visual condition planner computes a control weight for each input image based on the position of current sentence in the story. More precisely, the weight $\omega_j$ for image $I_j$ is:
\begin{equation}
    \omega_j = C\exp(-\frac{(i-j)^2}{2\sigma^2}),
\end{equation}
where $i \in \{1, \dots, N\}$ represents the position of current sentence in the story, and $C$ is a constant to normalize the weights and insure $\sum_{j=1}^N\omega_j = 1$. When $i=j$, the current sentence should be the exact description of image $I_j$, while remaining coherent to other images $I_{k\neq j}$.
Therefore, the weight of $I_j$ is the largest, and weight of $I_{k}$ descends as the distance $|k-j|$ grows.
Finally, the planner applies weighted multiplication on the scores of different images to obtain a unified matching score between the candidate sequence $x_{1:t}$ and the input images $\mathcal{I}$. Formally, 
\begin{equation}
    P_w(\mathcal{I}|x_t,x_{<t}) = \Pi_{j=1}^N\mathbf{D}(x_{1:t}, I_j)^{\omega_j}.
\end{equation}

It is worth noting that in our experiments, the aforementioned process is applied to a subset of the entire vocabulary, thereby reducing the computational cost of encoding and aligning candidate text. Specifically, we select the top $K$ tokens predicted by the language generator as the subset $V^K_{(t)}$. Moreover, to eliminate the bias of the cross-modality alignment results, we normalize the scores among candidate tokens. 
The final output of the visual condition planner can be written as:
\begin{equation}
    P(\mathcal{I}|x_t,x_{<t}) = \frac{e^{(P_w(\mathcal{I}|x_t,x_{<t}))}}{\sum_{x_i\in V^K_{(t)}}e^{(P_w(\mathcal{I}|x_i,x_{<t}))}}.
\end{equation}

\subsection{Visual-Conditioned generation}
Given the token probability predicted by the language generator and the aggregated cross-modality matching score, visual storytelling can be decoupled into the combination of language modeling and cross-modality aligning. Similar to Eq.~\eqref{Eq:decoupledprob}, the probability of next token $x_t$ can be decoupled as follows:
\begin{equation}
    \label{Eq:guidedgen}
    P(x_t|x_{<t}, \mathcal{I}) \propto P_{LM}(x_t|x_{<t})P(\mathcal{I}|x_t,x_{<t})^\gamma,
\end{equation}
where the hyper-parameter $\gamma$ controls the weight of visual information in the language generation process. While a higher value of $\gamma$ can improve the alignment of visual semantics, it may also adversely affect the quality of the generated language. Finding the right balance between language and visual information is crucial for achieving high-quality visual storytelling.

Furthermore, inspired by Su et al.~\cite{Su2022LanguageMC}, we incorporate a degeneration penalty into Eq.~\eqref{Eq:guidedgen} to prevent the repetitive degeneration problem. The final probability of visual-conditioned is formulated as:
\begin{equation}
    \label{Eq:finalgen}
    \begin{split}
    P(x_t|x_{<t}, \mathcal{I}) = &P_{LM}(x_t|x_{<t})P(\mathcal{I}|x_t,x_{<t})^\gamma -\\
    &\beta (\max(s(x_t, x_j), j\in \{1,\dots,t-1\}),
    \end{split}
\end{equation}
where $\beta$ is a hyper-parameter to control the degeneration penalty strength, and $s(x_i, x_j)$ is defined in Eq.~\eqref{Eq:token_sim}.

\section{Experiments}

\textbf{Dataset.} 
We make evaluation on the widely-used VIST benchmark~\cite{Huang2016VisualS} for visual storytelling. VIST contains \num[group-separator={,}]{210819} images from \num[group-separator={,}]{10117} Flickr albums. 
Each sample in VIST contains five images selected from an album, and a five-sentence story is annotated as ground truth. After excluding broken images, the dataset contains \num[group-separator={,}]{40098}, \num[group-separator={,}]{4988}, and \num[group-separator={,}]{5050} samples for training, validation, and testing respectively.
We use the test split of VIST as the evaluation benchmark in all experiments.
Following previous works~\cite{wang-etal-2018-metrics,9577852}, we evaluate at the album level, generating one story for each album regardless of different selected images.
During the training stage, we use the text part of the VIST training split, where all names are replaced with special placeholders.

\noindent\textbf{Implementation Details.} 
The language generator is initialized with a pre-trained GPT-2 model, and fine-tuning is performed on 2 GTX3090 GPUs for \num[group-separator={,}]{40000} steps with batch size of 256. We set the training loss weight $\alpha$ to 1.
To implement the visual discriminator, we ultilize a pretrained CLIP with ViT-base architecture as the image encoder.
The visual-conditioned generation is performed on 1 GTX3090 GPU.
In the reported results, we set the hyper-parameters $K$, $\gamma$, and $\beta$ to 45, 1, and 0.01, respectively.

\noindent\textbf{Evaluation Metrics.}
Following the existing works on the VIST benchmark, we adopt a set of automatic evaluation metrics including METEOR (M)~\cite{banerjee-lavie-2005-meteor}, BLEU (B-n)~\cite{papineni-etal-2002-bleu}, ROUGE\_L (R\_L)~\cite{lin-2004-rouge} and CIDEr (C)~\cite{7299087}. 
METEOR measures the semantic alignment between generated and reference sentences by leveraging WordNet.
BLEU computes the unigram and n-gram overlap between generated and candidate sentences. 
ROUGE\_L measures sentence-level similarity by computing the length of longest common subsequence.
CIDEr evaluates the consensus based on n-grams and weights n-grams using Term Frequency Inverse Document Frequency~(TF-IDF) to emphasize informative content.
However, we note that these metrics, as they rely on word correspondence with the ground truth, may not fully capture the quality of open-ended generation tasks such as storytelling.

\subsection{Quantitative Results}
\begin{table}[tb]
\renewcommand\arraystretch{1.2}
\begin{tabular}{ccccccc}
\hline

\multirow{2}{*}{Method} & \multicolumn{2}{c}{VIST-Text} & \multicolumn{2}{c}{ROC} & \multicolumn{2}{c}{WP} \\
\cmidrule(r){2-3} \cmidrule(r){4-5} \cmidrule(r){6-7}
                        & D-1 & D-2 & D-1 & D-2 & D-1 & D-2  \\
\hline
MAGIC & 2.4 & 6.6 & 0.8 & 1.3 & 1.1 & 2.5 \\
\textbf{Ours}  & \textbf{4.7} & \textbf{17.2} & \textbf{5.5} & \textbf{18.8} & \textbf{7.4} & \textbf{26.5}  \\
\hline
\end{tabular}
\caption{Diversity evaluation results. ``VIST-Text'', ``ROC'', ``WP'' represents the training of language generator is conducted on the text part of VIST dataset, ROCStories, and WritingPrompts, respectively. ``D-$n$'' refers to ``Distinct-$n$''. The best results for each metric are highlighted in bold.}
\label{Tab:disteval}
\vspace{-1em}
\end{table}


\textbf{Comparison with Existing Methods.}
We compare the generation quality of our method with a text-only trained methods. First, we adopt top-$k$ sampling~\cite{fan-etal-2018-hierarchical} ($k=40$) and nucleus sampling~\cite{Holtzman2019TheCC} ($p=0.95$). Since these sampling-based decoding strategy takes no account of visual inputs, we consider them as the lower bound of the text-only trained methods.
We also include MAGIC~\cite{Su2022LanguageMC}, which was proposed for image captioning and image-based story generation. MAGIC takes an image as input and generate text outputs by adding CLIP similarity scores on language model predicted probabilities. To extend MAGIC to the visual storytelling task, we average the representation of input image sequence to form the visual input of MAGIC.
To provide a comprehensive comparison, we also report results of several fully-supervised baselines.
INet~\cite{Jung2020HideandTellLT}, TAPM~\cite{9577852}, OIAVist~\cite{Braude2021OrderedAF}, and KAGS~\cite{9996128}

In Table~\ref{Tab:mainresult}, we present the comparison of our proposed method with existing fully-supervised and text-only trained methods.
As expected, the fully-supervised methods trained on cross-modality paired data exhibit better performance compared to the text-only trained methods.
However, our proposed method outperforms the text-only trained baselines on almost all metrics by a considerable margin, demonstrating the effectiveness of our visual-conditioned generation strategy.

\begin{figure}[tb]
    \centering 
    \subfloat{\includegraphics[width=0.5\linewidth]{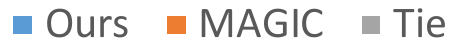}}
    
    \hfill
    \subfloat[Relevance]{\includegraphics[width=0.3\linewidth]{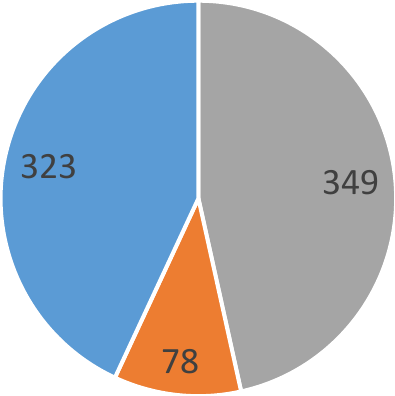}}
    \hfill
    \subfloat[Expressiveness]{\includegraphics[width=0.3\linewidth]{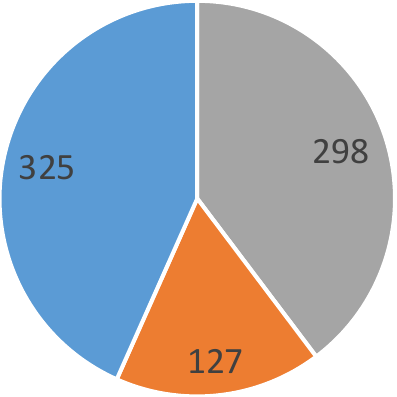}}
    \hfill
    \subfloat[Concreteness]{\includegraphics[width=0.3\linewidth]{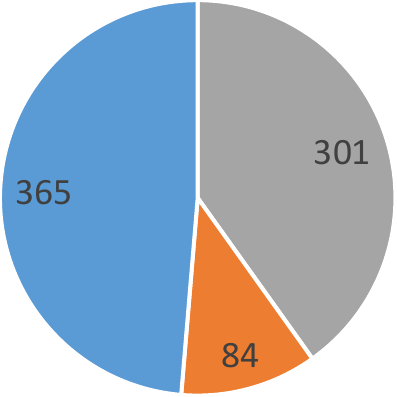}}
    \hfill

    \caption{Human evaluation results. ``Tie'' means the annotator cannot choose the better story.}
    \label{fig:HEeval}
\vspace{-1em}
\end{figure}

\begin{table}[tb]
\renewcommand\arraystretch{1.2}
\begin{tabular}{cccccccc}
\hline
Method                 & M    & B-1  & B-2  & B-3 & B-4 & R\_L & C   \\
\hline
Ours-Max               & 22.6 & 41.8 & 18.9 & 8.5 & 4.1 & 17.0 & 0.9 \\
Ours-Mean              & 22.8 & 43.2 & 20.0 & 9.1 & 4.4 & 17.2 & \textbf{1.3 }\\
Ours-Local             & 22.4 & 42.1 & 19.4 & 8.7 & 4.2 & 17.2 & 1.0 \\
\textbf{Ours-Planner } & \textbf{23.0} & \textbf{43.7} & \textbf{20.2 }& \textbf{9.2} & \textbf{4.5} & \textbf{17.3} & 1.2 \\
\hline
\end{tabular}
\caption{Evaluation results of different image album aggregation strategies. The best results for each metric are highlighted in bold.}
\label{Tab:ablation}
\vspace{-1em}
\end{table}

\noindent\textbf{Cross-domain Transfer.}
In order to evaluate the generalization ability of our method, we also explore cross-domain transfer by using story datasets of different domains in the text-only training stage. Specifically, we use
\textbf{ROCStories}~\cite{mostafazadeh-etal-2016-corpus} and \textbf{WritingPrompts}~\cite{fan-etal-2018-hierarchical} for training.
The training split of ROCStories dataset contains \num[group-separator={,}]{51165} five-sentence commonsense stories.
And the training split of WritingPrompts dataset contains \num[group-separator={,}]{272600} stories collected from Reddit’s WRITINGPROMPTS forum\footnote{www.reddit.com/r/WritingPrompts/}. The average length of WritingPrompts stories is 734.5, and the average number of sentences is 39.4, making it significantly larger than the VIST dataset and introducing a larger domain gap. 
During training, we exclude the story title and writing prompts to align with the VIST evaluation process.

In Table~\ref{Tab:domsintrans}, we compare the cross-domain transfer ability  between our method and the text-only trained baselines. We observe a considerable drop in performance for all methods when evaluated on datasets from different domains. This is expected since the style, theme and topic of the stories are different across datasets. However, our method still outperforms others on most evaluation metrics, demonstrating its superior generalization ability.

\noindent\textbf{Diversity Evaluation.}
To further assess the expressive diversity of the generated stories, we use Distinct-$n$ which calculates the number of distinct n-grams of all generated stories~\cite{li-etal-2016-diversity}. The value is divided by the total number of generated tokens to avoid favoring long sentences.
The results presented in Table~\ref{Tab:disteval} demonstrate that our method significantly outperforms the baseline in terms of diversity. This can be attributed to the ability of our method to attend to both global and local visual input, which results in more informative and diverse expressions.
Additionally, it can be observed that the diversity of generated stories is relevant to the training corpus, which suggests that the incorporation of external text corpus can benefit visual storytelling.

\begin{figure}[tb]
    \centering 
    \includegraphics[scale=0.32]{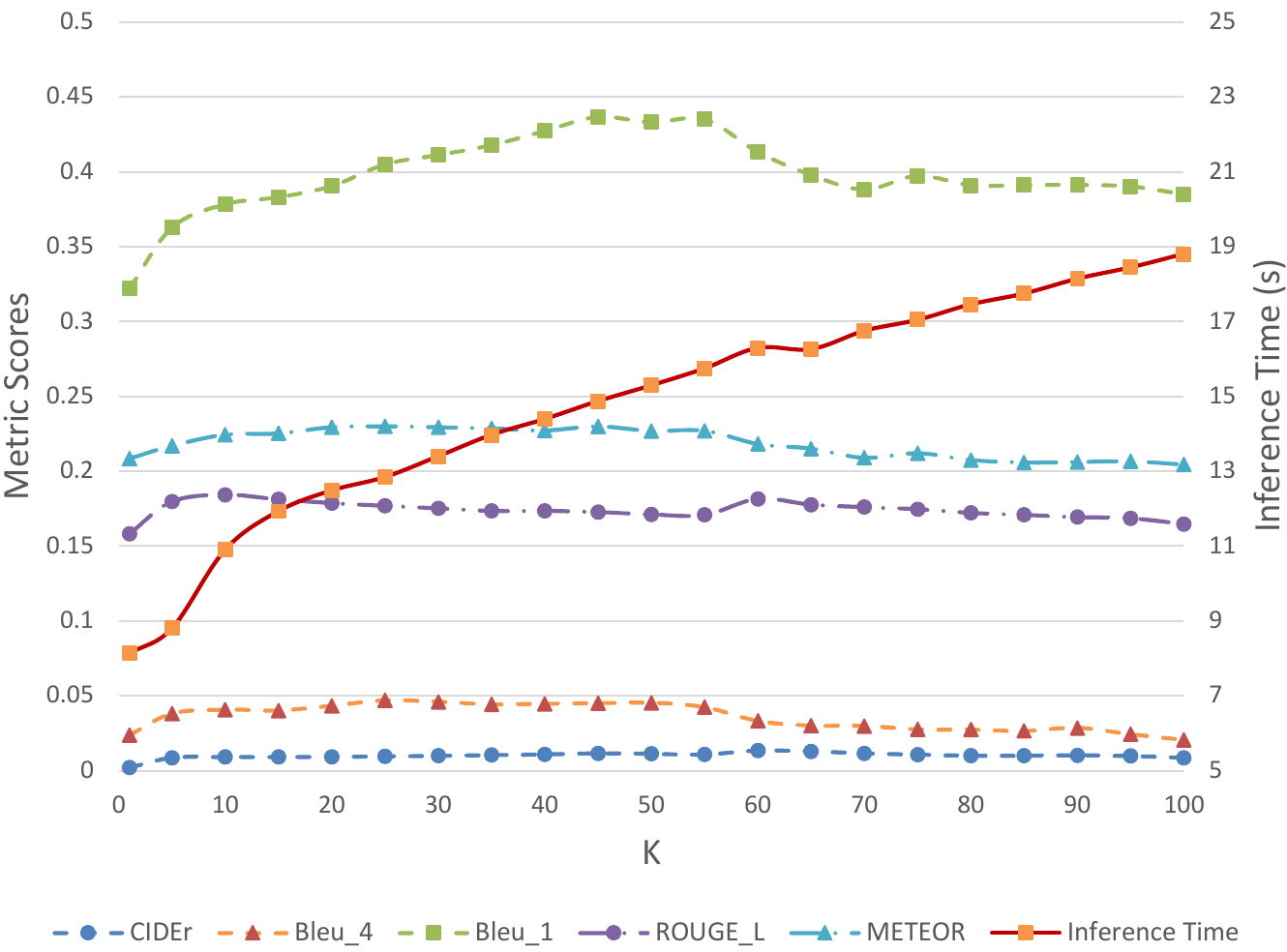}
    \caption{Analysis of the effect of number of candidates $K$.}
    \label{fig:K_ablation}
\vspace{-1.5em}
\end{figure}

\noindent\textbf{Human Evaluation.}
As illustrated in previous works~\cite{wang-etal-2018-metrics,Braude2021OrderedAF}, automatic evaluation metrics are insufficient for visual storytelling due to its subjective and imaginative nature.
To obtain more reliable estimates, we also perform human evaluation. 
Following common practice, we randomly selected 150 examples from the test set, and invited 5 human annotators to rank generation results of different methods. 
Specifically, the annotators were asked to evaluate the stories based on three criteria: relevance, expressiveness, and concreteness. Relevance refers to whether the story covers the topic and main objects in the images. Expressiveness refers to whether the story is coherent, grammatically and semantically correct, and free of repetition. Concreteness refers to whether the story is narrative and concrete.

Fig.~\ref{fig:HEeval} shows the evaluation results of 5 human annotators. Our method outperforms MAGIC by a large margin in all three aspects. 
The dominance of our method is most significant in terms of Concreteness, indicating a greater ability to incorporate visual details in the generated stories.
The expressiveness of MAGIC is better than the two other aspects, which reflects the fact that the language quality of our method is slightly affected by introducing fine-level visual control.
Additionally, the ``Tie'' option is selected in a large percentage in all three criteria, which has not been reported in previous methods~\cite{Jung2020HideandTellLT,9577852,9996128,Braude2021OrderedAF}. We believe the reason is that the overall quality of stories generated by text-only trained methods is lower than full-supervised methods, making it difficult to rank for human annotators.

\subsection{Ablation Study}

\begin{figure}[tb]
    \centering 
    \includegraphics[scale=0.29]{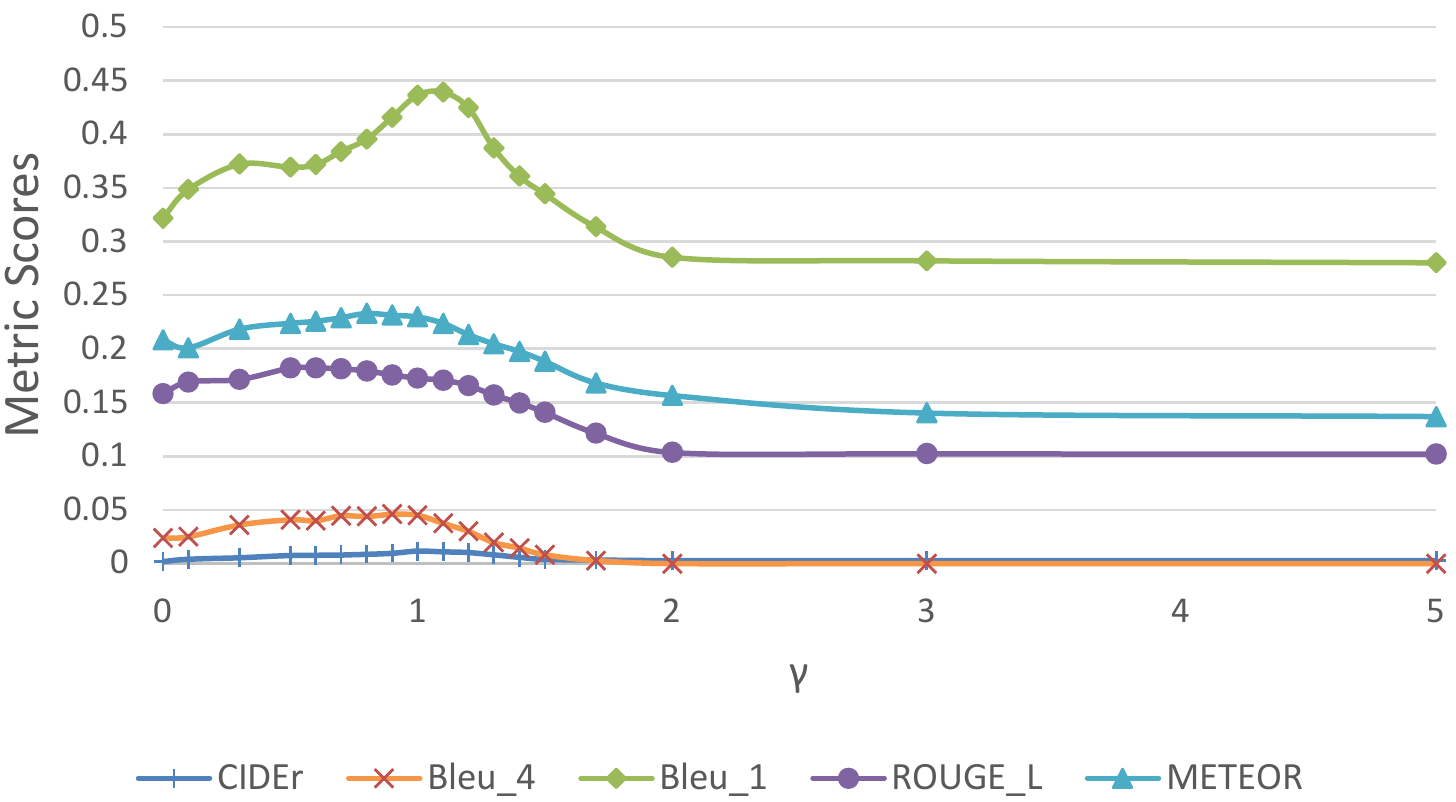}
    \caption{Analysis of the effect of control weight $\gamma$.}
    \label{fig:controlweight_ablation}
\vspace{-1.5em}
\end{figure}

\begin{figure*}[tb]
    \centering
    \includegraphics[scale=0.50]{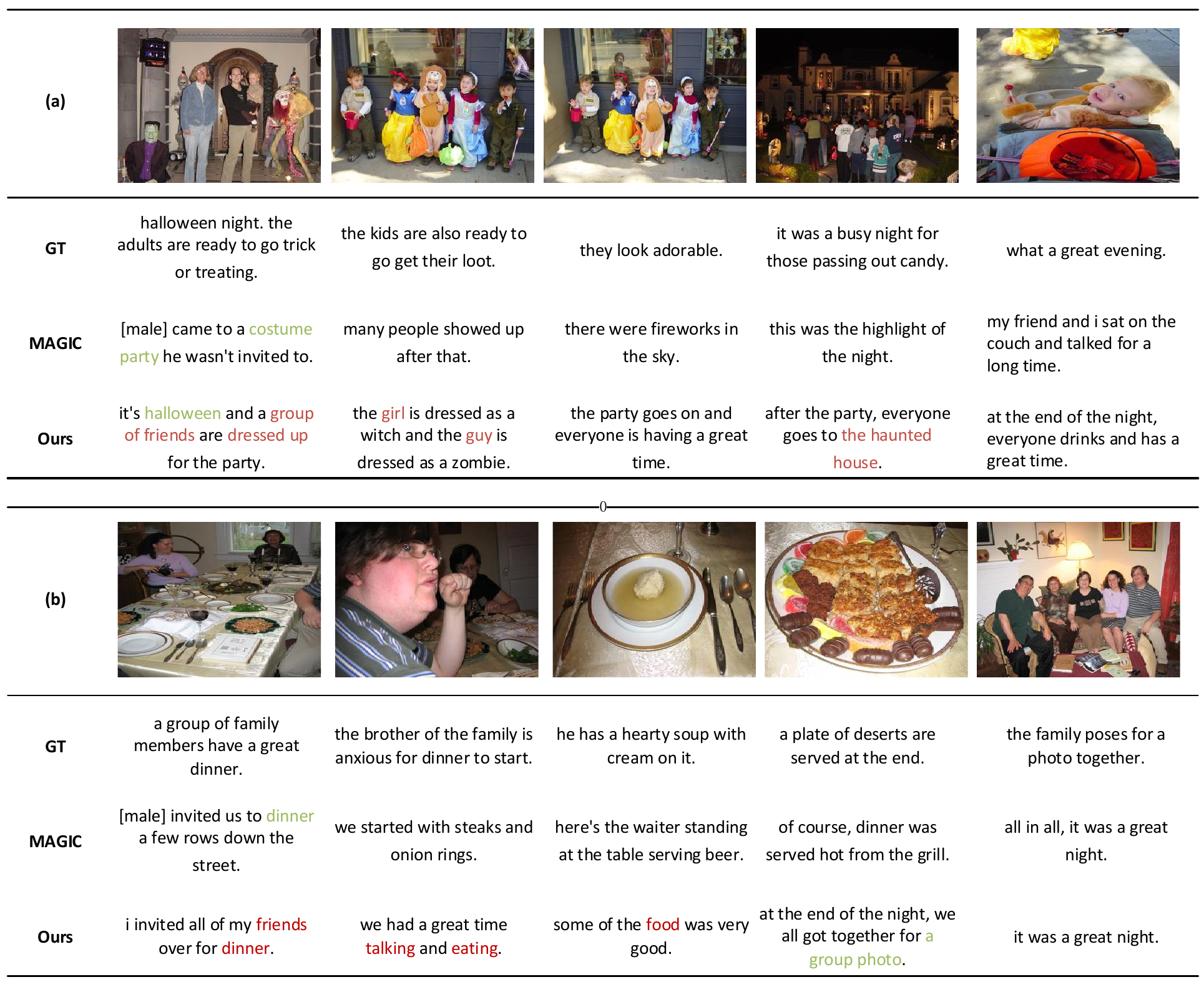}
    \caption{Qualitative comparison between our method and MAGIC. Words highlighted in \textcolor[RGB]{192,0,0}{red} represents exact description of corresponding image, and words highlighted in \textcolor[RGB]{157,187,97}{green} represents information from other images.}
    \label{fig:qual_eval}

\end{figure*}
\textbf{Impact of Visual Condition Planner.}
We conduct ablation experiments to analyze the effect of the visual condition planner, which aggregates the cross-modality matching result of input images.  Specifically, we replace the aggregation process to three straightforward strategies: 1) choosing the maximum matching score in all images, 2) averaging the scores of all images, and 3) using the score of the corresponding image. 
The evaluation results in Table~\ref{Tab:ablation} indicate that both strategies of viewing the images equally within the album (``Ours-Max'' and ``Ours-Mean'') and focusing solely on the corresponding local image (``Ours-Local'') have a negative impact on the quality of the generated stories.

\noindent\textbf{Impact of Hyper-parameters.}
During the visual-conditioned generation, the selection of top-$K$ candidate tokens to compute cross-modality matching score with visual inputs and the addition of visual control to the decoding process with a control weight $\gamma$ in Eq.~\eqref{Eq:finalgen} are governed by predefined hyper-parameters. Therefore, it is important to analyze the influence of these hyper-parameters on the quality of generated stories.

From the results in Fig.~\ref{fig:K_ablation}, we observe that the performance improves with $K$ when $K<30$, and remains relatively stable when $K$ continues to increase. However, when $K$ is too large ($>60$), the performance slightly decreases as $K$ keep increasing. 
It is also worth noting that the inference time significantly increases as $K$ increases.
Therefore, we choose $K=45$ in our experiments as it strikes a balance between performance and efficiency.
The results in Fig.~\ref{fig:controlweight_ablation} demonstrate the significant impact of the control weight $\gamma$ on the generation process.  Specifically, when the control weight is too small, the generated stories tend to be disconnected from the visual input, while an excessively large control weight will lead to a disruption in the decoding process, thus deteriorating the overall quality of the generated text. These experimental findings align with our initial intuition and suggest the importance of selecting an appropriate control weight in the visual-conditioned generation.

\subsection{Qualitative Results}
Fig.~\ref{fig:qual_eval} presents two examples of generated stories by MAGIC and our proposed method. 
The results show that our approach generates stories with more accurate semantics that correspond to the images, as indicated by the red highlights. 
Moreover, the visual condition planner enables the generation of sentences that are relevant to the other images in the input sequence, as shown by the green highlights. 
Our method outperforms the baseline method in capturing visual contents within a single image and maintaining the theme of the album, resulting in stories of higher quality.


\section{Conclusion and Discussion}
In this paper,  we propose a novel approach for visual storytelling that only requires textual story data for training. By leveraging the capabilities of pretrained cross-modality models such as CLIP, we model the visual storytelling task as a visual-conditioned generation problem. We adopt a guided decoding paradigm and design a visual condition planner to aggregate the input visual sequence. Our method is evaluated on the VIST benchmark through extensive experiments, which demonstrate its effectiveness in generating high-quality visual stories.

Although the proposed method avert the cost of cross-modality annotated data, the training-free visual condition planner does have its limitations in understanding the complex temporal structures of visual input, which may affect the complexity of the generated story. 
In future work, it may be worth exploring few-shot learning methods to aggregate aligning results of image sequence to generate more informative and narrative stories.

\begin{acks}
This work was supported by NSFC under Contract U20A20183 and 62021001. It was also supported by the GPU cluster built by MCC Lab of Information Science and Technology Institution, USTC, and the Supercomputing Center of the USTC.
\end{acks}

\clearpage
\bibliographystyle{unsrt}
\bibliographystyle{ACM-Reference-Format}
\bibliography{sample-base}

\end{document}